\documentclass[letterpaper, 10 pt, conference]{ieeeconf} 
\IEEEoverridecommandlockouts                             
\usepackage{color}
\usepackage{color,soul}

\usepackage{graphicx} % for pdf, bitmapped graphics files
\usepackage{mathtools} 
\usepackage{siunitx}
\usepackage[table]{xcolor}
\usepackage{cite}

\usepackage{graphicx}
\usepackage{setspace}
\usepackage{acronym}
\usepackage{amsfonts}			% special math characters
\usepackage{times}		% use Postscript fonts
\usepackage{color,soul}
\usepackage{amssymb}
\usepackage{amsmath}
\usepackage{gensymb}
\usepackage{tabularx} % in the preamble
\usepackage{multirow}
\usepackage{algorithm}
\usepackage{algorithmic}
\usepackage{booktabs}
\usepackage[T1]{fontenc}

\usepackage{float}
\usepackage{mwe}
\newcolumntype{L}{>{\centering\arraybackslash}m{3cm}}
\usepackage[thinc]{esdiff}
\usepackage[table]{xcolor}
\usepackage{hyperref}
\hypersetup{
    colorlinks=true,
    citecolor=black,
    linkcolor=black,
    filecolor=black,      
    urlcolor=black,
    }
\usepackage{url}
\usepackage{array,ragged2e}
\newcolumntype{P}[1]{>{\RaggedRight\arraybackslash}p{#1}}

\newcolumntype{M}[1]{>{\centering\arraybackslash}m{#1}}

\usepackage[font=small,skip=2pt]{caption}
\usepackage{textcomp, gensymb}
\usepackage{xcolor}
\usepackage{etoolbox}
\usepackage{caption}
\usepackage{multirow}
\usepackage{booktabs}
\usepackage{subcaption}
\newbool{showcolor}
\setbool{showcolor}{true}  % Change to false to disable coloring

\newcommand{\changed}[1]{%
  \ifbool{showcolor}{\textcolor{blue}{#1}}{#1}%
}

\newcommand{\changedmore}[1]{%
  \ifbool{showcolor}{\textcolor{teal}{#1}}{#1}%
}

\usepackage{microtype}
\renewcommand{\baselinestretch}{1}

\title{\LARGE \bf On Steerability Factors for Growing Vine Robots}
\author{Ciera McFarland$^{1*}$, Antonio Alvarez Valdivia$^{2*}$, Sarah Taher$^{1}$, Nathaniel Hanson$^{2\dag}$, and Margaret McGuinness$^{1\dag}$
\thanks{Correspondence: {\tt\small nhanson2@mit.edu}}
\thanks{$^{*}$Equal Contribution, $^{\dag}$Equal Contribution}
\thanks{$^{1}$University of Notre Dame, Notre Dame, Indiana, USA}
\thanks{$^{2}$Lincoln Laboratory, Massachusetts Institute of Technology, Lexington, Massachusetts, USA}
\thanks{
DISTRIBUTION STATEMENT A. Approved for public release. Distribution is unlimited.
This material is based upon work supported by the Department of the Air Force under Air Force Contract No. FA8702-15-D-0001 or FA8702-25-D-B002. Any opinions, findings, conclusions or recommendations expressed in this material are those of the author(s) and do not necessarily reflect the views of the Department of the Air Force. © 2025 Massachusetts Institute of Technology. Delivered to the U.S. Government with Unlimited Rights, as defined in DFARS Part 252.227-7013 or 7014 (Feb 2014). Notwithstanding any copyright notice, U.S. Government rights in this work are defined by DFARS 252.227-7013 or DFARS 252.227-7014 as detailed above. Use of this work other than as specifically authorized by the U.S. Government may violate any copyrights that exist in this work.}}

\begin{document}

\maketitle
\thispagestyle{plain}
\pagestyle{plain}

\begin{abstract}
Vine robots extend their tubular bodies by everting material from the tip, enabling navigation in complex environments. Despite their promise for field applications, performance is constrained by the weight of attached sensors, and design and control choices. This work investigates how tip load, pressure, length, diameter, and fabrication method shape vine robot steerability--the ability to maneuver with controlled curvature--for robots that steer with series pouch motor-style pneumatic actuators. We conduct two groups of experiments: (1) studying tip load, chamber pressure, length, and diameter in a robot supporting itself against gravity, and (2) studying fabrication method and ratio of actuator to chamber pressure in a robot supported on the ground. Results show that steerability decreases with increasing tip load, is best at moderate chamber pressure, increases with length, and is largely unaffected by diameter. Robots with actuators attached on their exterior begin curving at low pressure ratios, but curvature saturates at high pressure ratios; those with actuators integrated into the robot body require higher pressure ratios to begin curving but achieve higher curvature overall. We demonstrate that robots optimized with these principles outperform those with \textit{ad hoc} parameters in a mobility task that involves maximizing upward and horizontal curvatures.
% Vine robots extend their tubular bodies by everting material from the tip, enabling navigation in complex environments with a minimalist soft body. Despite their promise for field applications, especially in the urban search and rescue domain, performance is constrained by the weight of attached sensors or tools, as well as other design and control choices. This work investigates how tip load, pressure, length, diameter, and fabrication method shape vine robot steerability—the ability to maneuver with controlled curvature--for robots that steer with series pouch motor-style pneumatic actuators. We conduct two groups of experiments: (1) studying tip load, chamber pressure, length, and diameter in a robot supporting itself against gravity, and (2) studying fabrication method and ratio of actuator to chamber pressure in a robot supported on the ground. Results show that steerability decreases with increasing tip load, is best at moderate chamber pressure, increases with length, and is largely unaffected by diameter. Robots with actuators attached on their exterior begin curving at low pressure ratios, but curvature saturates at high pressure ratios; those with actuators integrated into the robot body require higher pressure ratios to begin curving but achieve higher curvature overall. We demonstrate that robots optimized with these principles outperform those with \textit{ad hoc} parameters in a mobility task that involves maximizing upward and horizontal curvatures.
\end{abstract}

\section{Introduction}

Robots are routinely used to gather important information from dangerous environments; to be effective, they must move through these environments while carrying the sensors and tools needed for their task. Vine robots, soft inflatable robots that extend by everting their body from their tip \cite{blumenschein2020design}, have shown promise in navigating confined spaces. They have previously explored urban search and rescue training sites \cite{der2021roboa, mcfarlandSPROUT}, an archaeological site \cite{CoadRAM2020}, and a salamander burrow habitat \cite{qin20253d}. They are also capable of squeezing through small apertures \cite{HawkesScienceRobotics2017} and growing vertically~\cite{CoadRAM2020}. However, these demonstrations largely relied on \textit{ad hoc} design and control parameters, limiting their generalization.

Vine robot mobility is constrained when they carry tip-mounted loads \cite{mcfarlandSPROUT} or must span open spaces \cite{mcfarland2023collapse}. Because tip mounts, sensors, and tools add mass while altering structural and control demands, it is essential to systematically quantify how design and control parameters—such as tip weight, internal pressures, length, diameter, and body fabrication—influence steerability. \textit{Steerability}, defined here as the ability of a vine robot to generate controlled curvature in free space, is critical for maneuvering in complex environments and for ensuring reliable field performance.

\begin{figure}[t]
    \centering
    \includegraphics[width = \columnwidth]{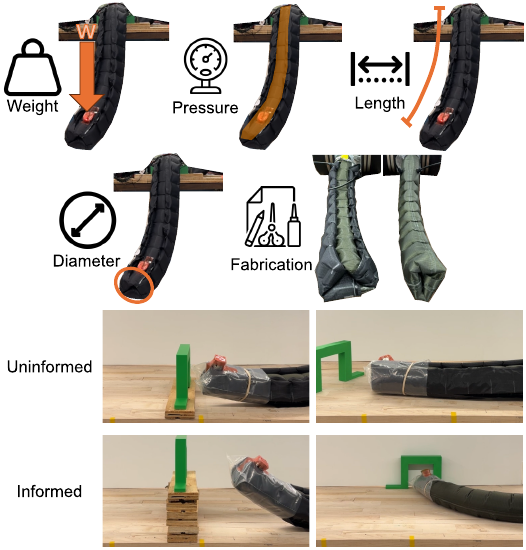}
    \caption{\textbf{Recommendations for robot operation.} Our research shows that steerability is higher when a robot has lower tip weight, a moderate chamber pressure relative to its actuators, and a longer length. Diameter has minimal impact, while the ideal fabrication method depends on what pressure ratios the robot will operate at. A robot with uninformed parameters steers to lower curvatures than a robot with informed parameters.}
    \label{glamor shot}
    \vspace{-2.0em}
\end{figure}

This work addresses this gap by experimentally characterizing how vine robot steerability depends on design and control parameters (Fig. \ref{glamor shot}). Specifically, we focus on vine robots that steer using series pouch motor-style pneumatic actuation~\cite{CoadRAM2020, Ataka2020ModelBased}. We examine (1) how tip load, chamber pressure, length, and diameter affect a robot supporting itself against gravity; and (2) how actuator placement (exterior \cite{mcfarland2023collapse} vs. integrated~\cite{Ataka2020ModelBased}) and the ratio of actuator to chamber pressure affect planar curvature in robots supported on the ground. While prior studies have characterized unweighted curvature~\cite{kubler2024comparison}, growth speed~\cite{alvarez2025high}, and steering force~\cite{Mendoza2024HighCurvature} for vine robots, our focus is identifying trends in 3D steerability under loads, enabling the development of generalizable design and control principles for pneumatically steered vine robots.

% The goal of this work is to determine how well a vine robot can steer in a vertical plane when loaded under different weights, operating at different pressures and lengths, and when the robot is built to different diameters in free space. We also explore how steerability depends on body fabrication methods from the literature (exterior vs. integrated actuators) and the ratio of actuator pressure to chamber pressure, which together determine how actuation forces translate into curvature. While other studies have focused on achieving the highest possible curvature or high growth speed and force \cite{alvarez2025high}, this particular work is meant to explore trends within the steerability space to help improve the performance of any vine robot. This analysis will be used to provide general recommendations for vine robot design and control parameters. These tests are conducted with a vine robot design that has been validated under extreme field conditions \cite{mcfarlandSPROUT,CoadRAM2020},  but we also evaluate a vine robot with embedded actuators to more thoroughly replicate the scope of vine robots in the literature.

The specific contributions of this work are as follows:
\begin{itemize}
    \item A systematic analysis of how tip load, chamber pressure, length, and diameter affect vine robot steerability under self-supporting 3D conditions.
    \item A comparative study of exterior vs. integrated actuator designs and how actuator fabrication and pressure ratios influence steerability under supported planar conditions.
    \item A demonstration that robots designed with these principles achieve superior task performance compared to robots with \textit{ad hoc} parameters.
\end{itemize}

\section{Related Work}

Continuum robot research has shown that where and how actuation is applied strongly influences deformation and workspace \cite{russo2023continuum, seleem2023recent}. For vine robots, the choice of steering mechanism has a significant impact on the curvature the robot can achieve. Discrete methods such as latches~\cite{HawkesScienceRobotics2017} or locally placed magnets~\cite{li2021bioinspired} provide sharp curvatures at predefined locations, but are challenging to reverse. Internal tip devices~\cite{der2021roboa, lee2023high} allow high-curvature steering of only the robot tip, which is less useful for navigating free space. External magnets~\cite{kim2025external} can steer vine robots, but only in small-scale environments where magnetic forces can reach. 

Various pneumatic actuators offer reversible, continuous curvature steering for vine robots, including series pneumatic artificial muscles~\cite{greer2017series}, fabric pneumatic artificial muscles~\cite{Naclerio2020simple}, and cylindrical pneumatic artificial muscles~\cite{kubler2023multi}. Series pouch motors~\cite{CoadRAM2020, Ataka2020ModelBased} are another type of pneumatic actuator that are easily fabricated using impulse heat-sealing or heat pressing~\cite{mack2025efficient}. These actuators can be interfaced with the main vine robot main chamber tube either as external elements attached to the body wall~\cite{CoadRAM2020} or integrated within the wall itself~\cite{Ataka2020ModelBased}. The quantitative impact of fabrication methods on free-space steerability has not been studied.

Other parameters relevant to free-space steerability include tip load, internal pressure, and body geometry. Vine robots must balance the need to carry sensors and tools at the tip with the risk of collapse under transverse loading or axial buckling. Increasing a vine robot's internal pressure or diameter~\cite{mcfarland2023collapse}, or incorporating stiffening elements~\cite{do2024stiffness} helps it better support its own weight but also affects its ability to steer. Previous studies have looked at how vine robots steer when using obstacles to guide their path on flat ground~\cite{GreerICRA2018, selvaggio2020obstacle} and have compared curvatures that can be achieved with various pneumatic actuators~\cite{kubler2024comparison} without tip loading. No study has systematically compared design and control parameters for 3D mobility tasks in free space for series pouch motor-actuated robots. We expect our results to directly inform other pneumatically steered vine robots, and to provide helpful insights for vine robots steered through other means.

% Prior work has identified many factors that influence vine robot steering, including payload weight, chamber pressure, body length, and geometry \cite{heap2024large, CoadRAM2020, HawkesScienceRobotics2017, greer2019soft}. However, these effects have largely been reported qualitatively through demonstrations or application-focused studies rather than systematic analysis. Likewise, while fabrication methods have been mentioned as practical considerations in vine robot steering, their quantitative impact on free-space steerability remains unclear. This motivates our study, which provides a controlled comparison of traditional design parameters and fabrication styles with the goal of developing general design and control recommendations.

\section{Self Supporting 3D Steerability Experiments} \label{sec:parameter analysis}
In this section, we present the setup and results of four sets of experiments that evaluate steerability trends for vine robots with exterior pouches; these experiments vary tip load, chamber pressure, length, and diameter.

\subsection{Robot Body Fabrication}
For all experiments in this paper, we use robot bodies made from 30 denier heat-sealable thermoplastic polyurethane (TPU)-coated ripstop nylon fabric (extremtextil, Dresden, Germany), with a main chamber and three actuators spaced circumferentially around the robot body and attached along the robot length. In this section, we use robots constructed with \textit{exterior} pouches, described in Sec.~\ref{sec:planarsteerability}-A. 

\subsection{Experimental Setup}

\begin{figure}[tb]
\centering
\includegraphics[width = \columnwidth]{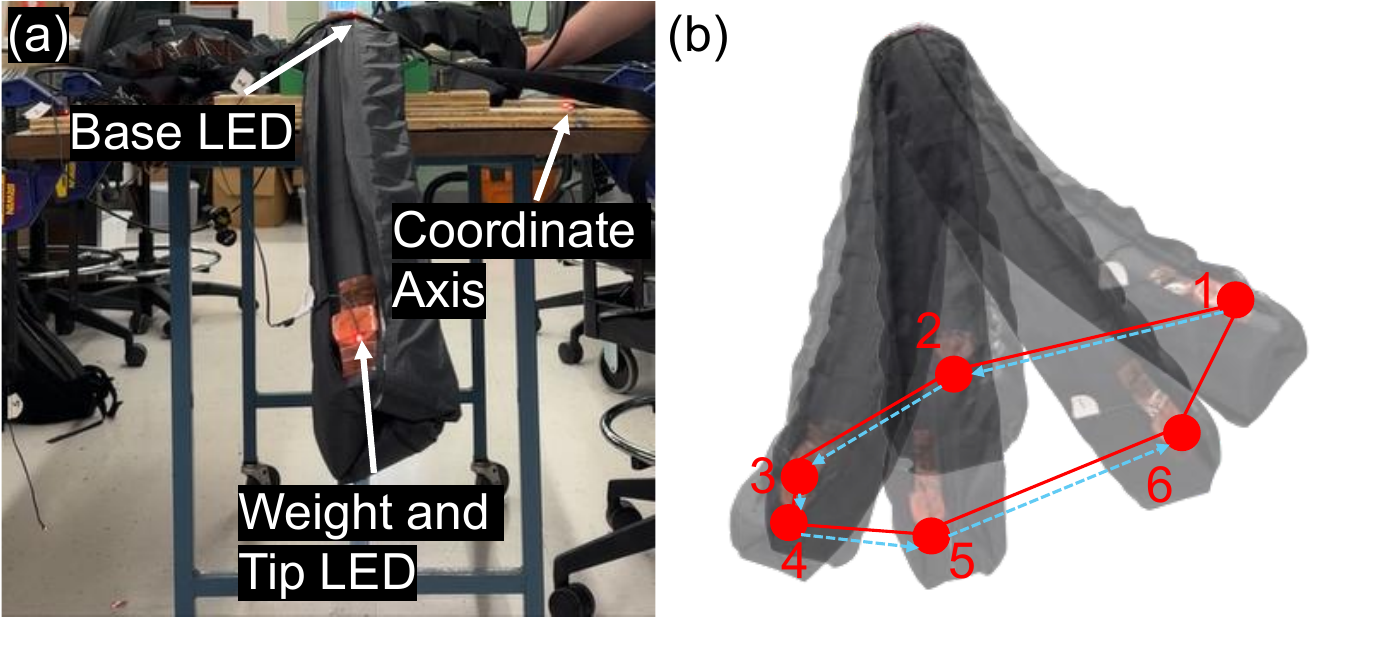}
\caption{\textbf{Experimental setup and points of interest for self-supporting 3D steerability experiments.} (a) The robot is placed on a table with support only underneath the base section. Tracking the base LED, tip LED, and coordinate axis allows us to align the data with a reference frame fixed to the base of the robot. (b) The order in which the robot tip visits the six via points is shown in blue. The convex hull area of those points is shown in red.}
\label{workspace setup}
\vspace{-2.0em}
\end{figure}

\begin{figure*}[tb]
\centering
\includegraphics[width=0.95\textwidth]{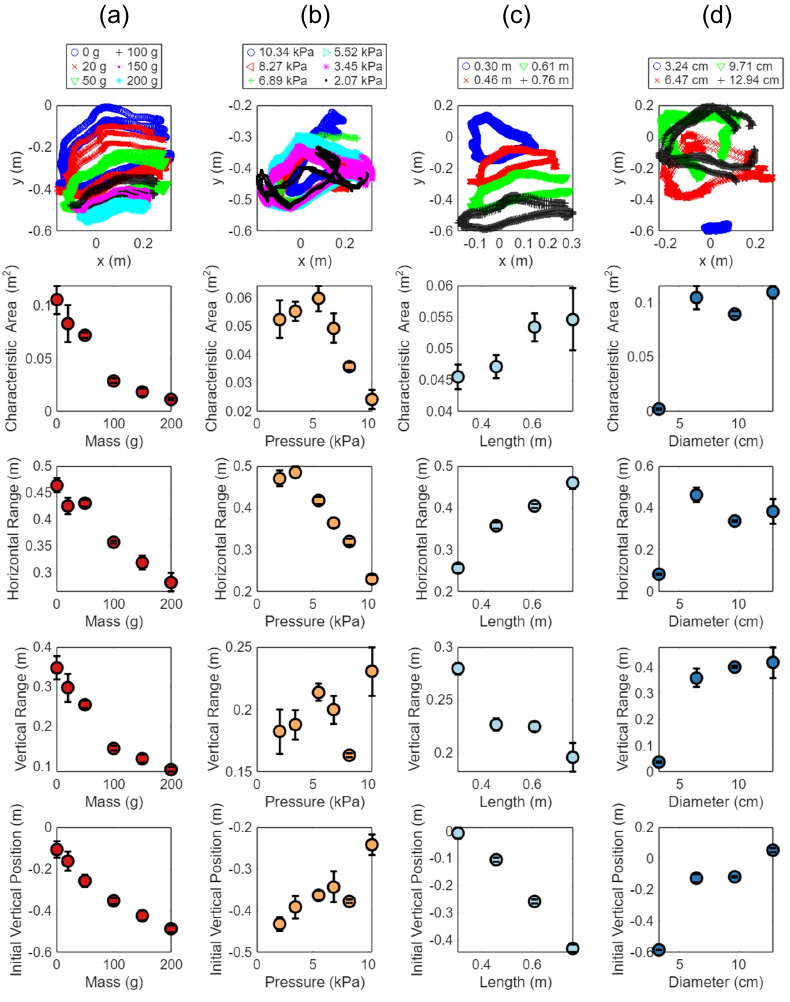}
\caption{\textbf{Results of self-supporting 3D steerability experiments}. For all parameters, we plot the paths traced out by the robot tip during each trial, and the average and standard deviation for the characteristic area, horizontal range, vertical range, and initial vertical position. (a) Increasing tip load significantly decreases characteristic length, horizontal range, vertical range, and initial vertical position. (b) Increasing pressure initially increases and then decreases characteristic length; it significantly decreases horizontal range and changes vertical range erratically. The initial vertical position generally increases. (c) Increasing robot length increases characteristic length and significantly increases horizontal range, while decreasing vertical range and initial vertical position. (d) As long as the robot is not collapsed (leftmost data point), varying diameter does not significantly impact characteristic length, horizontal range, vertical range, or initial vertical position.}
\label{key experiments}
\vspace{-2.0em}
\end{figure*}

The experimental setup is shown in Fig.~\ref{workspace setup}(a). For each set of trials, the robot had a fixed tip load, chamber pressure, length, and inflated diameter of the main chamber tube. To prevent the robot body from changing length, we heat sealed the inner material (the ``tail'') to the outer material (the ``wall'') at the robot base and clamped it together. We mechanically constrained the robot to be cantilevered horizontally off a table with a fixed distal point of support while it moved its tip in free space towards maximum curvature in every direction. 

For each trial, the robot moved through six via points (red dots connected by a blue path in Fig.~\ref{workspace setup}(b)), chosen to represent the full range of motion of the robot in the two degrees of freedom of its 3D tip position that are controlled by the steering actuators. We inflated each actuator in turn to its maximum pressure of 10.34~kPa (determined using a safe margin from the pressure at which we expected the heat seals to tear), as well as each pairing of two actuators to their maximum pressure at the same time, with the other actuators set to zero pressure. We used four closed-loop pressure regulators (QB3, Proportion-Air, McCordsville, IN) to enforce the desired values for chamber and actuator pressures. Before commanding the next set of pressures, we waited for the robot to stop moving. We conducted three trials for each set of parameters. The chamber was deflated between trials, and the robot was reset to have all actuators at zero pressure, to ensure a more repeatable path. 

We used a motion capture (Impulse X2E, PhaseSpace, San Leandro, CA) LED marker to track the 3D tip position of the robot at a rate of 30~Hz. This tip-mounted LED is attached to the top of any weights on the robot on the top edge of the robot. We also used LEDs aligned with our desired coordinate frame and an LED on the top edge of the base of the robot to realign the data to our global reference frame located at the base of the robot with the $z$-axis pointing along the robot toward its tip and the $y$-axis pointing up. 

To visualize the steerability of the robot, we project the 3D tip position data into the $x$-$y$ plane and calculate the convex hull area of the region enclosed by the $x$ and $y$ coordinates of $m$ marker positions (red shape in Fig.~\ref{workspace setup}(b)) using Gauss's area formula. We describe this as our \textit{characteristic area}:
\begin{equation}
    \label{eq:characteristic_length}
    A_{c} =\tfrac{1}{2}\left|\sum_{k=1}^{m}(x_k y_{k+1}-x_{k+1} y_k)\right|,
\end{equation}
where $x_{m+1} = x_1$ and $y_{m+1} = y_1$. 

We chose to analyze this characteristic area as the metric for the steerability of the robot under the various conditions tested, as opposed to either a traditional 3D workspace or a curvature measurement. A 3D workspace of the robot would plot all reachable robot tip positions, considering growth (one degree of freedom) and steering (two degrees of freedom). As a proxy for this 3D workspace measurement that allows us to examine trends as various parameters change, the characteristic area metric can be thought of as the area of a vertical slice through the 3D workspace of the robot, which represents the steerability of the robot at a given length. 
% This allows us to measure two dimensions of the  workspace at several discrete lengths, focusing on understanding how the steerability changes with the different parameters tested. 
A curvature measurement (as we used in the supported planar steerability experiments described in Section~\ref{sec:planarsteerability}) does not capture the robot's behavior well when moving unsupported in 3D space, because the robot's body deflects under the influence of gravity, not forming a constant curvature shape. Additionally, the robot's range of motion is different in the horizontal and vertical directions, due to differing effects of gravity, so measuring the horizontal and vertical range of the positions reached at each constant length allows us to quantify that effect.
% which would include $z$ position data, due to the unique characteristics of the robot. This $x$-$y$ plane projection shows in which directions the robot is able to orient itself and ultimately grow. The $z$ position data would only provide information regarding how much the robot would need to grow in order to reach a given location. This is not of interest for steerability, as shown in Figure~\ref{workspace setup}(c). Additionally, we chose not to examine curvature for these self-supporting 3D steerability experiments . While curvature provides strong information regarding how much the robot can turn in a given direction, it does not provide the full picture of how much the robot can curve in every direction in free space for any configuration. Therefore, we chose to judge steerability using our characteristic area metric.  

Fig.~\ref{key experiments} shows the results of our self-supporting 3D steerability experiments. For each set of parameters tested (columns (a) through (d)), we plot the paths traced out by the robot tip during each trial (top row). We also plot the average characteristic area across three trials and its standard deviation (second row), along with the average and standard deviation for the \textit{horizontal range} (the difference between the minimum and maximum $x$ values) (third row), the \textit{vertical range} (the difference between the minimum and maximum $y$ values) (fourth row), and the \textit{initial vertical position} (the $y$ value at the start of a trial, which begins when the robot reaches via point 1) for each trial (fifth row).  

\subsection{Tip Load}
% Vine robots are strong in tension, but weak when experiencing compressive or transverse forces. While tip-mounted tools are critical for field work, they are also detrimental to a vine robot's functionality due to their weight. Therefore, it is critical to understand what loads the robot bears reasonably while still functioning to a desirable degree.

When varying the tip load, we maintained the following constant parameters: a chamber pressure of 6.89~kPa, which is high enough to support the robot while also being low enough for it to steer freely; a length of 0.61~m, which is sufficiently long to steer without causing the robot to excessively deflect under its own weight; and an inflated diameter of 8.1~cm, which is close to the size of previously deployed field robots~\cite{CoadRAM2020, der2021roboa, mcfarland2023collapse}. The robot performed its task under loads of 0~g, 20~g, 50~g, 100~g, 150~g, and 200~g (the maximum mass the robot could hold before it started approaching collapse behavior).

Steerability as a function of tip load is shown in Fig.~\ref{key experiments}(a). There is a clear decrease in characteristic area with increased tip load, with a total decrease of 89\% occurring between a load of 0~g and a load of 200~g. Correspondingly, the horizontal range, vertical range, and initial vertical position decrease by 39\%, 74\%, and 0.38~m, respectively.

\subsection{Chamber Pressure}
% A vine robot varies its stiffness by changing the pressure in its main chamber. High pressures are associated with increased stiffness, which is useful for supporting the robot's weight against gravity as well as achieving faster growth. Increased stiffness is expected to also decrease steerability. Identifying what pressures support the robot while still allowing it to steer as needed for a given task is important to determining its operating parameters.

When varying chamber pressure, we maintained the following constant parameters: a tip load of 50~g, an intermediate load that causes some deflection but not so little steerability as to prevent the analysis of other parameters; a length of 0.61~m; and a diameter of 8.1~cm. The robot performed its task at chamber pressures of 2.07~kPa (close to the minimum chamber pressure that supports the robot), 3.45~kPa, 5.52~kPa, 6.89~kPa, 8.27~kPa, and 10.34~kPa (close to the maximum chamber pressure it can hold without bursting).

Steerability as a function of chamber pressure is shown in Fig.~\ref{key experiments}(b). The characteristic area starts high at a low pressure and initially increases 14\% with pressure, peaking at 5.52~kPa. Beyond this point, higher pressures lead to a 54\% decrease in characteristic area compared to the initial value, as the robot is too stiff for the actuators to move it effectively. The horizontal range initially increases 3\%, but ultimately decreases by 51\% with increasing pressure, which is expected with the robot's increasing stiffness. The vertical range generally increases around 26\% with increasing pressure but experiences high variation. This is likely due to the fact that the robot is becoming stiffer and harder to move, but that same stiffness is also preventing deflection and therefore allowing the robot to make larger motions if it creates enough force to do so. Similarly, we see the initial vertical position increase 0.19~m over the pressure range, as a stiffer robot can start each trial with a higher tip position.

\subsection{Length}
% Vine robots must grow in order to navigate a space. Changes in length are critical to their nature. Therefore, it is important to know how length changes affect steerability, as this information could be used for path planning purposes. 

When varying robot length, we maintained the following constant parameters: a tip load of 50~g; a chamber pressure of 6.89~kPa; and a diameter of 8.1~cm. The robot performed its task at lengths of 0.30~m, 0.46~m, 0.61~m, and 0.76~m. 

Steerability as a function of robot length is shown in Fig.~\ref{key experiments}(c). Generally, the characteristic area increases with increasing length, with a 20\% increase in characteristic area occurring from 0.30~m to 0.76~m. There is also an 80\% increase in horizontal range, which makes sense, since a longer robot's tip should travel farther for a given amount of curvature. However, the vertical range, which decreases by 30\%, shows that this increase in length results in a heavier robot that struggles more to move against gravity. Eventually, the robot would become so long that it would collapse, reducing the characteristic area, horizontal range, and vertical range to nearly zero. We also see the initial vertical position dropping by 0.42~m, significantly limiting where in the vertical plane we can reach.

\subsection{Diameter}

When varying diameter, we maintained the following constant parameters: a tip load of 50~g; a chamber pressure of 6.89~kPa; and a length of 0.61~m. The robot performed its task at inflated diameters of 3.2~cm (smaller than the typical field exploration robot), 6.5~cm, 9.7~cm, and 12.9~cm (larger than the typical field exploration robot). It should be noted that the 9.7~cm diameter robot was oriented upside down for its trials, and therefore it formed a differently-oriented trajectory and had a correspondingly lower initial vertical position.

Steerability as a function of robot diameter is shown in Fig.~\ref{key experiments}(d). There is only a 4\% change in characteristic area between the 6.5~cm diameter robot and the 12.9~cm diameter robot. The 3.2~cm robot collapsed under the given parameters. Thus, having a diameter large enough to avoid collapse is important, but how much larger is of minimal consequence. Similarly, the horizontal range, vertical range, and initial vertical position are fairly constant once the robot is at a sufficient diameter to not collapse, though there is a 17\% decrease in horizontal range, a 16\% increase in vertical range, and a 0.18~m increase in initial vertical position with increasing diameter, excluding the 3.2~cm robot.

% \subsection{Body Construction}
% Here we describe the conceptual differences in the soft growing body models we are experimenting with: exterior pouches and integrated pouches. We show a diagram of the concepts, and maybe explain how they are constructed. 

\section{Supported Planar Steerability Experiments}
\label{sec:planarsteerability}
In this section, we present the setup and results of experiments that evaluate the effect of the fabrication method used to incorporate series pouch motor actuators, and the pressure ratio between the actuators and the robot's main chamber, on steerability. Fabrication determines the location of actuator forces relative to the body wall, while the pressure ratio governs how those forces interact with the body stiffness.

\subsection{Robot Body Fabrication}
\begin{figure}[t]
\centering
\includegraphics[width = 0.95\columnwidth]{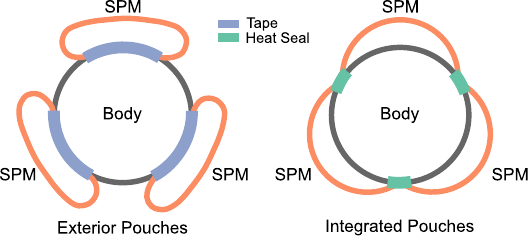}
\caption{\textbf{Cross-section diagram of robot body fabrication methods.} (a) \textit{Exterior} pouches: series pouch motor actuators are heat sealed separately and taped to the outside of the main chamber tube. (b) \textit{Integrated} pouches: pouch chambers are heat sealed directly to the main chamber tube. Both designs use thermoplastic polyurethane-coated ripstop nylon.}
\label{fig:body_construction}
\vspace{-2.0em}
\end{figure}

We compare two robot body fabrication approaches (Fig.~\ref{fig:body_construction}): \textit{exterior} and \textit{integrated} pouches. In this section, we use robots with main chamber tube inflated diameter of 7.6~cm.

In the \textit{exterior} pouch design~\cite{CoadRAM2020}, the main chamber is first fabricated as a single-layer thermoplastic polyurethane (TPU)-coated ripstop nylon tube. Three series pouch motor actuators, each with half the lay-flat width of the main chamber tube, and with pouches of equal length and width (experimentally determined to yield maximum contraction~\cite{kubler2024comparison}) are then constructed and taped outside the main chamber. 
% Because the actuators apply force directly on the outer wall, we expect this design to produce noticeable steering at relatively low pouch motor pressures. However, the workspace may be constrained by the fact that actuation forces are applied only at the body surface.

In the \textit{integrated} pouch design~\cite{Ataka2020ModelBased}, the main chamber and actuators are built together as a double-layer structure. A folded piece of TPU-coated fabric is heat-sealed to define both the body wall and internal pouch chambers before being joined lengthwise into a tube. In this construction, each series pouch motor actuator has a nominal width of two-thirds of the lay-flat diameter of the main chamber tube.
% Because the coated sides of the fabric are internal to the layering, eversion is smoother, and internal friction during growth is reduced. 
% Here, actuation forces are embedded within the structure itself, but steering is likely to require higher pressure ratios before significant bending occurs

\subsection{Pressure Ratios}
% In addition to fabrication method and body structure, we theorize steering performance depends strongly on the relative pressures applied to the vine robot’s body chamber and its pouch actuators. 
To compare pressures across body designs, we define the pressure ratio as:
\begin{equation}
        \text{ratio} = \frac{P_{pouch}}{P_{body}},
\end{equation}
where $P_{pouch}$ is the pressure supplied to the series pouch motor actuator to achieve maximum curvature and $P_{body}$ is the chamber pressure maintaining the body inflated. This ratio normalizes actuation input relative to body support, providing a consistent framework to examine how steering forces are transmitted through the structure. Low ratios ($<1$) correspond to cases where chamber pressure dominates, maintaining stiffness but potentially limiting curvature. High ratios ($>1$) may increase bending but also reduce the robot's stability. 
% By systematically varying the ratio, we evaluate how each fabrication method converts actuation input into steering motion.
\label{sec:pr}
\subsection{Experimental Setup}
\begin{figure}[t]
\centering
\includegraphics[width = \columnwidth]{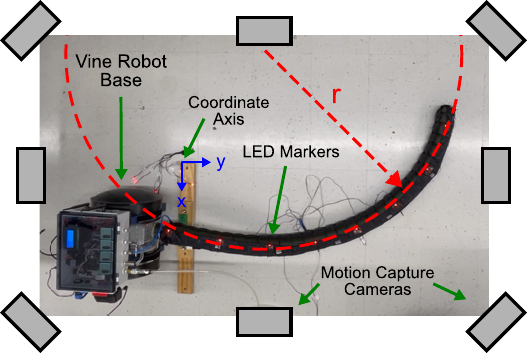}
\caption{\textbf{Experimental setup for supported planar steerability experiments.} The robot (1.75~m long) is entirely supported on the floor as it curves due to changing actuator and chamber pressure. Motion capture markers are placed along the centerline at 25 cm intervals to capture the robot shape and tip position.}
\label{fig:curvature_setup}
\vspace{-1.0em}
\end{figure}

We conducted a series of planar bending trials with both body types (Fig.~\ref{fig:curvature_setup}). The robots were held at a constant length of 1.75~m using a motorized spool within a vine robot base similar to the one in~\cite{CoadRAM2020}. Pressure was controlled using the same pressure regulators as in Sec.~\ref{sec:parameter analysis}-B. For each trial, a single actuator was aligned with the measurement plane, isolating its contribution to steering. Unlike the free-space trials in Sec.~\ref{sec:parameter analysis}, these experiments were performed with the robot lying on the floor, which reduced the influence of body weight, length, and gravitational loading along the body, allowing more direct characterization of fabrication methods and pressure ratio alone.

The robot body was tracked using motion capture LED markers  with the same system as in Sec.~\ref{sec:parameter analysis}-B,  placed along the centerline at 25 cm spacing, starting at the base. This enabled reconstruction of the planar backbone for consistent extraction of the tip position and curvature. Similar to the previous experiments, we also placed LED markers aligned with the desired coordinate frame, this time with $y$ horizontal along the robot's axis from base to tip when straight and $z$ up, and an LED at the base of the robot.

Pouch and chamber pressures were varied independently between 3.45 and 13.79~kPa, in 3.45~kPa increments, resulting in pressure ratios ranging from 0.25 to 4.0. Because all combinations of inputs were tested, some ratios corresponded to multiple chamber–pouch pairs (e.g., two combinations at ratio $= 2$). For each set of parameters, three trials were performed. A trial began with the main chamber inflated to the desired baseline pressure while all pouch motors were unpressurized. A single pouch motor line was then actuated to the commanded pressure, and motion capture was used to record the full sequence of steering, including the initial straight state, the buildup of curvature, and the final state after the pouch was deflated. After each trial, the robot was fully deflated, manually straightened to the zero state, and reinflated to the baseline chamber pressure before beginning the next trial. Steering was quantified by fitting a circle to the planar tip trajectory using least squares to compute curvature as $\kappa = 1 / r$, where $r$ is the circle's radius. 

\begin{figure}[t]
\centering
\includegraphics[width = 1.0\columnwidth]{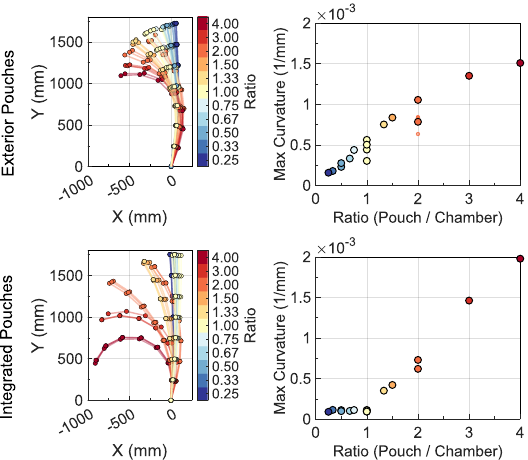}
\caption{\textbf{Results of supported planar steerability experiments.} Plots show (left) robot shapes and (right) corresponding maximum curvatures for vine robots with (top row) exterior and (bottom row) integrated pouch actuators across a range of pressure ratios ($P_{pouch}/P_{body}$). In the plots on the right, small markers indicate results of individual trials, while large markers denote the mean of three trials for each condition. The exterior pouch robot begins curving at lower pressure
ratios, but its curvature begins to saturate at high pressure ratios. The integrated pouch robot requires higher pressure
ratios to begin curving but achieves higher curvature overall.}
\label{fig:curvature_summary}
\vspace{-2.0em}
\end{figure}

\begin{figure}[t]
\centering
\includegraphics[width = \columnwidth]{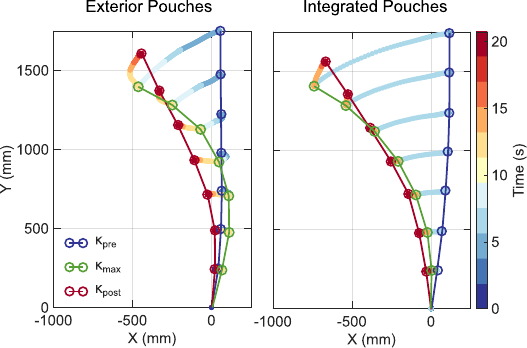}
\caption{\textbf{Example steering trajectories while actuator pouch is inflated and then deflated.} Plots show time lapse shapes for (left) exterior and (right) integrated fabrications at a constant $P_{body} =$ 6.89~kPa and a maximum $P_{pouch} =$ 13.78~kPa (ratio $= 2$). Color bar indicates time progression from chamber inflation ($t=0$~s) through pouch inflation to maximum pressure (green markers) to pouch deflation to zero pressure ($t=20$~s). In both fabrications, the robot does not reset to its initial configuration after pouch deflation, demonstrating hysteresis and residual deformation.}
\label{fig:curvature_example}
\vspace{-2.0em}
\end{figure}

\subsection{Results} \label{sec:curvature_results}
Fig.~\ref{fig:curvature_summary} summarizes the influence of body fabrication and pressure ratio on vine robot steering. Both body types exhibit increasing curvature as pouch pressure rises relative to body pressure (higher ratios). However, the extent of the accessible curvatures and the rate at which curvature increases differ substantially between the fabrications. With the exterior pouches, bending initiates at lower ratios, but the maximum curvature achieved across the tested conditions is notably smaller. With the integrated pouches, no curvature is observed until the ratio exceeds $1$. Beyond this threshold, curvature increases approximately linearly with pressure ratio and reaches higher overall values. 

The observed differences in the robot's steerability based on the fabrication method arise from hoop stress effects. When the body chamber is pressurized, the wall is placed in hoop tension. Integrated pouches share this stressed wall, so their “bulging” strains must be generated against an existing preload, delaying the onset of curvature until $P_{pouch}$ sufficiently exceeds $P_{body}$ (no bending until $P_{pouch} / P_{body}>1$). Exterior pouches, in contrast, are mechanically decoupled from the body wall; they expand outward without overcoming hoop preload, which allows a bending moment about the neutral axis to be generated at lower ratios. However, as $P_{body}$ increases, the wall still resists in-plane shortening, so exterior pouch fabrications saturate earlier in workspace compared to integrated ones.

Fig.~\ref{fig:curvature_example} illustrates the steering dynamics for a representative case with $P_{body} =$ 6.89~kPa and $P_{pouch} =$ 13.79~kPa (ratio $=2$). The color bar shows the time course from initial chamber pressurization ($t=0$~s), through pouch inflation to maximum curvature (green connected markers), and subsequent deflation ($t=20$~s). In both fabrications, the robot does not fully return to its initial straight configuration. 
% This non-resetting behavior reflects hysteresis and residual deformation in the body wall, consistent with material compliance and inter-layer friction. 

% Overall, these results indicate that fabrication method and pressure ratio interact strongly to shape steering performance. Exterior pouch fabrication enables immediate, low-ratio steering, but with a constrained workspace, while integrated fabrication supports greater workspace exploration at the expense of requiring higher actuation pressures.

\section{Discussion}
% In the experiments outlined in Sec.~\ref{sec:parameter analysis}, the robot performs the commanded trajectory with the tail pinned. This is necessary to keep the length constant during the trials. When the tail was restricted by a motor, the force created by steering was sufficiently strong to pull on the motor and allow the robot to grow as trials continued. If a backward force was applied to the tail to prevent any motion, the force was too strong and resulted in the robot exhibiting shape-locking behavior. These shapes were often hard to repeat as the excess force did not result in the robot bending exactly the same way each time.

% Pinning the tail was a compromise between the two options available to a motorized robot, but it is not practical to use a vine robot in this manner in the field as growth is one of its primary advantages. Therefore, vine robots in the field may be growing or exhibiting some degree of shape locking while steering in response to the motor commands. The pinned tail robot was able to display relatively repeatable behavior, which is why it was the method we ultimately chose to draw general conclusions about how various parameters affect performance. 

\subsection{Position Resetting}
In Sec.~\ref{sec:curvature_results}, we observed that neither exterior nor integrated pouch fabrications returned to their initial straight configuration after inflating and then deflating the actuator (Fig.~\ref{fig:curvature_example}). This non-resetting reflects hysteresis and residual deformation of the body wall, likely due to material compliance and inter-layer friction. 
% For these experimental trials, each repetition ended with the robot in its deformed state, and the chamber and actuator were then deflated, manually straightened, and reinflated before beginning the next trial to ensure consistency. 
% Although this method allowed for systematic testing, it also highlights a limitation: 
Without manual intervention, the vine robots we studied here accumulate residual curvature, which reduces repeatability. For practical use, this behavior not only makes control strategies that assume a perfect reset unreliable, but also constrains the robot’s effective steerability in 3D space. 
% In field settings, this could limit the ability to precisely direct the tip to new targets after multiple maneuvers. 
This motivates the need for future research on robot body designs that exhibit less hysteresis in steering, and/or that use proprioceptive sensing to detect and compensate for residual curvature.
% , enabling operators to maintain awareness of the robot’s true shape and tip position.
% This is a general issue for vine robots, not one unique to our setup, and addressing it will be critical for achieving robust and repeatable operation in the field.
%Vine robot implementations may need either (i) active reset mechanisms, such as reverse actuation or chamber venting routines, or (ii) calibration procedures that account for residual offsets between trials.

\subsection{Design and Control Recommendations}
When looking at the parameters that affect steerability, it is important to note that only some of them are modifiable in the field. The robot's diameter and fabrication method will be fixed from the start. Length will vary as the robot moves through the space by nature of its operation. A multi-sensor payload will ideally not have a variable weight, but sensors could be removed to improve the robot's steerability in the field. The most practical parameters to directly modify mid-task are the pressures applied to the chamber and pouches.

\subsubsection{Diameter}

% The diameter of a vine robot should be chosen such that it will not collapse under the desired operating tip load, pressure, and length in free space. Its minimal impact on steerability suggests it should be largely based on the needs of the given environment. 

Robot diameter primarily acts as a feasibility constraint rather than a continuous tuning parameter for steerability. In our experiments, the smallest tested diameter (3.2~cm) collapsed under a 50~g tip load at 6.89~kPa and a free length of 0.61~m, resulting in near-zero characteristic area (Fig.~\ref{key experiments}(d)). In contrast, all larger diameters (6.5 -- 12.9~cm) remained stable and produced similar characteristic areas, horizontal ranges, vertical ranges, and initial vertical positions. These results indicate that once a vine robot's diameter exceeds the minimum required to avoid collapse for a given load, pressure, and length, increasing diameter primarily rescales the resulting geometry rather than introducing new steering capability \cite{blumenschein2018helical}. As a result, diameter selection should prioritize environmental constraints such as aperture size or available deployment volume, rather than steering performance.  

\subsubsection{Fabrication Method}

% Our results show that fabrication fundamentally shapes steering. Exterior pouch fabrication allows bending to initiate at relatively low ratios, enabling immediate responsiveness but limiting the total accessible workspace. Integrated pouch fabrication requires the ratio to exceed $1$ before any curvature is observed, but once this threshold is reached, curvature increases approximately linearly with the ratio, and the robot is enabled to access a larger workspace.

% Whether the exterior pouch fabrication or the integrated pouch fabrication should be chosen for a particular task depends on at what pressure ratio the task will be able to be completed. For tasks that must be completed at high pressure ratios (usually due to needing a high chamber pressure to achieve growth or prevent robot collapse), the exterior pouch design should yield higher curvature. For tasks that can be completed at high pressure ratios (i.e., ones where the robot body is supported and/or not actively growing while steering), the integrated pouch design should yield higher curvature. 

The choice between exterior and integrated pouch fabrications determines both the pressure ratio required to initiate steering and the maximum curvature that can be achieved. As shown in Fig.~\ref{fig:curvature_summary}, exterior pouch robots begin to curve at low pressure ratios ($P_{pouch}/P_{body} < 1$), but their curvature saturates at higher ratios, limiting the maximum achievable bending. In contrast, integrated pouch robots exhibit no measurable curvature until the pressure ratio exceeds unity, after which curvature increases approximately linearly with pressure ratio and reaches higher peak values. These behaviors define two distinct operating regimes. Tasks that require steering while maintaining a high chamber pressure, such as growth under load or avoiding collapse in free space, benefit from exterior pouch fabrications, which can generate curvature at low ratios. Conversely, tasks performed in supported or quasi-static conditions, where higher pressure ratios are feasible, benefit from integrated pouch fabrications, which more effectively convert actuator pressure into large curvature once the hoop-stress threshold is exceeded.

\subsubsection{Length} 
% As long as the robot is not long enough to collapse, the robot's free length should be maximized. While the free length of the robot during exploration of a space depends largely on the shape of the space being explored, an informed operator can choose which path to control the robot to take. For example, given multiple options of apertures to travel through before executing a sharp turn, an operator might choose the aperture that happens earlier in the growth process so that the robot is left with a longer length to curve to achieve the turn.

Robot length introduces a trade-off between lateral reach and vertical maneuverability. In our experiments, increasing free length consistently expanded the characteristic area and horizontal range, as a longer body amplifies tip displacement for a given curvature (Fig.~\ref{key experiments}(c)). At the same time, vertical range and initial vertical position decreased with length, indicating a growing influence of gravity that limits upward steering even before full collapse occurs. 

As a result, increasing free length is beneficial for tasks dominated by horizontal steering, but can be detrimental for tasks that require lifting the tip or maintaining elevation in free space. While the robot’s length is inherently dictated by growth, an informed operator can exploit this trade-off through path planning. For example, when multiple apertures are available prior to a sharp turn, selecting an earlier aperture preserves greater usable length for lateral curvature while avoiding excessive gravitational penalty.
 
\subsubsection{Tip Load}

% The total weight of the sensor payload on the robot's tip should be as light as possible to minimize the impact on steerability. However, making certain sensors removable may also aid the robot in performing more difficult maneuvers in situations where not all sensors are needed.

Tip load directly constrains free-space steerability by amplifying gravitational and transverse forces at the robot tip. In our experiments, increasing payload mass produced a nonlinear reduction in steerability, with characteristic area, horizontal range, and vertical range all degrading sharply beyond approximately 100~g (Fig.~\ref{key experiments}(a)). Above this load, the robot approached collapse and exhibited severely limited directional reach. These results indicate that tip load is not merely a performance modifier, but a dominant limiting factor for 3D steering. While minimizing payload mass is always beneficial, selectively removing or disabling non-essential sensors during challenging maneuvers offers a practical means of restoring steerability without redesigning the robot body.

\subsubsection{Operating Pressures and Ratios}

% In supported planar tasks, the ratio of actuator to chamber pressure should be maximized. In self-supporting 3D tasks, a moderate chamber pressure should be selected to balance the tradeoff between the robot body being too soft to support itself against gravity and too stiff to steer. 

Operating pressures govern a fundamental trade-off between body stiffness, growth capability, and steering authority. In free-space 3D tasks, absolute chamber pressure exhibits a non-monotonic effect on steerability: low pressures provide insufficient support against gravity, while high pressures overly stiffen the body and limit actuator effectiveness, as reflected in both characteristic area and initial vertical position. In our experiments, characteristic area peaked at an intermediate chamber pressure, indicating an optimal stiffness range for unsupported steering. In supported planar tasks, steering performance is instead dominated by the relative pressure ratio between the pouch actuators and the body chamber. For both fabrication methods, curvature increased with $P_{pouch}/P_{body}$, but with distinct regimes: exterior pouch designs initiate steering at low ratios and saturate early, while integrated pouch designs require ratios exceeding unity before curving but achieve higher curvature at larger ratios.

Across both classes of tasks, feasible operating pressures are constrained by growth and structural limits. Chamber pressure must exceed the minimum required to sustain growth at the desired speed~\cite{HawkesScienceRobotics2017}, while both chamber and actuator pressures must remain below their respective burst limits. Improved robot fabrication techniques that allow higher chamber and actuator pressures are particularly promising, as they would expand the achievable overlap between high growth speed and high curvature, especially for integrated pouch designs.
% Rather than prescribing a single set of pressures, our results suggest a control paradigm in which pouch pressure serves as the primary steering input. For exterior pouch fabrications, steering is achieved across a wide range of ratios with relatively low pouch pressures. For integrated fabrications, chamber pressure must be kept low enough—or pouch pressure high enough—to exceed the ratio threshold and achieve curvature. In practice, operators may need to adjust both pouch and body pressures depending on the fabrication method. Because vine robots must balance steering with continued eversion, pressure management becomes a coupled problem. 
% At low chamber pressures, growth may stall entirely, restricting the ability to maintain a desired ratio. Conversely, increasing chamber pressure to sustain growth reduces the effective ratio below the threshold needed for integrated pouch designs to bend. 
Given current fabrication limitations, growth may need to pause temporarily while the robot steers into position, before reinflating the body chamber to resume extension. Control strategies that coordinate growth and steering -- for example, alternating between ``growth phases'' and ``steering phases'' -- could expand the range of environments where vine robots are deployed.

\section{Demonstration}

\begin{figure}[tb]
\centering
\vspace{1.0em}
\includegraphics[width = 0.95\columnwidth]{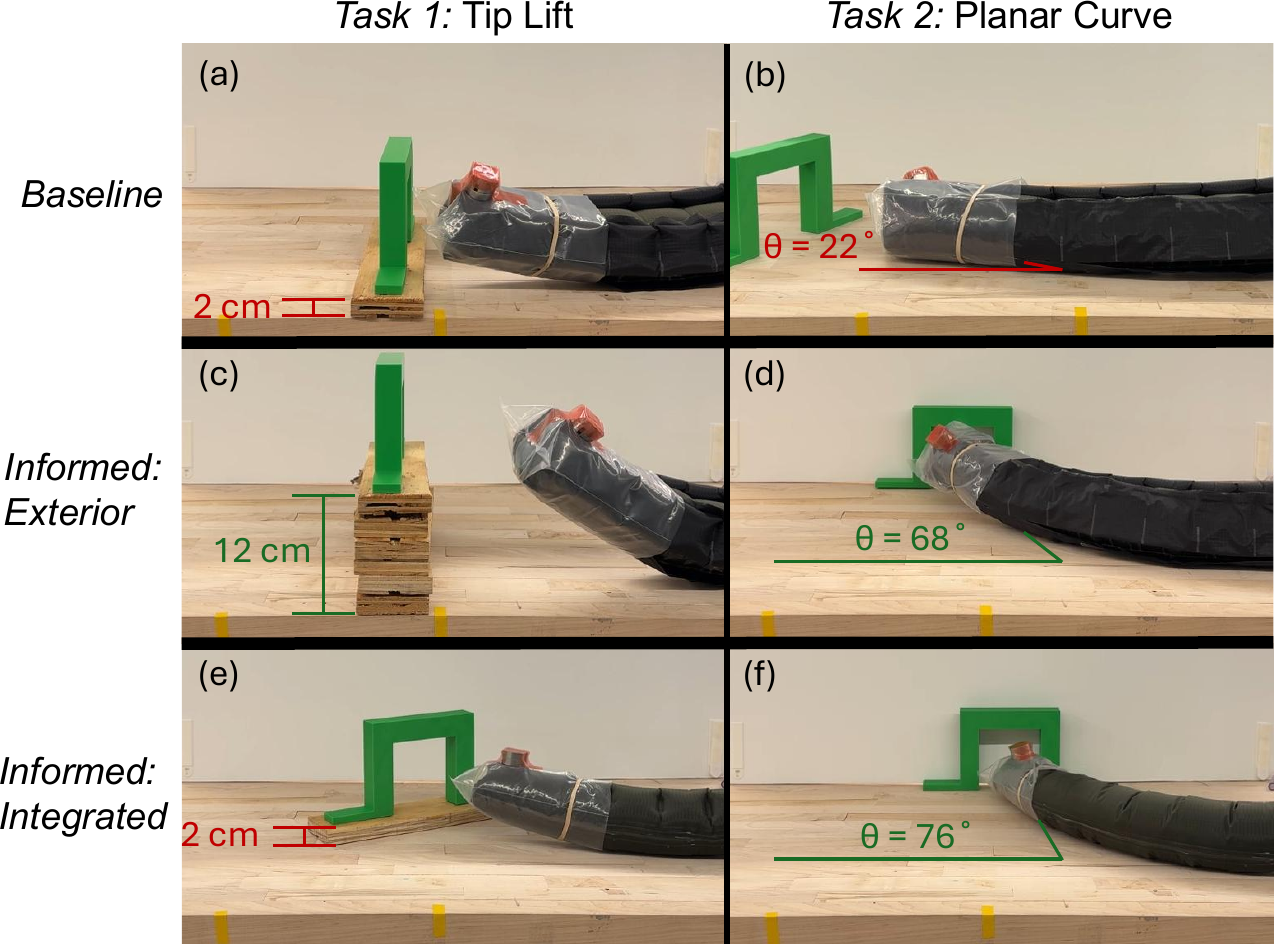}
\caption{\textbf{Demonstration of effect of tip weight, pressures, and fabrication method on mobility.} A robot with \textit{ad hoc} design and control parameters struggles to (a) lift its tip and (b) create a planar curve. A similarly-fabricated robot with deliberately-chosen parameters achieves (c) greater height and (d) a sharper planar curve. A robot with a different fabrication method achieves (e) similar height to the first robot, but (f) better planar curvature than either of the other designs. Parameters are listed in Table~\ref{table demo}.}
\label{demo}
\vspace{-2.0em}
\end{figure}

\begin{table*}[tb]
\centering
\vspace{0.5ex}
\begin{tabular}{lcccccc}
\toprule
 & \multicolumn{2}{c}{Design 1} & \multicolumn{2}{c}{Design 2} & \multicolumn{2}{c}{Design 3} \\
\cmidrule(lr){2-3} \cmidrule(lr){4-5} \cmidrule(lr){6-7}
 & Task 1 (a) & Task 2 (b) & Task 1 (c) & Task 2 (d) & Task 1 (e) & Task 2 (f) \\
\midrule
Chamber Pressure (kPa)   & 10.34 & 10.34 & 6.89 & 3.45 & 6.89 & 3.45 \\
Actuator Pressure (kPa)  & 10.34 & 10.34 & 13.79 & 13.79 & 13.79 & 13.79 \\
Tip Load (g)             & 150   & 150   & 100   & 100   & 100   & 100 \\
Fabrication              & Exterior & Exterior & Exterior & Exterior & Integrated & Integrated \\
Achieved Height (cm)     & 2     & n/a   & 12    & n/a   & 2     & n/a \\
Achieved Curve (°)       & n/a   & 22     & n/a   & 68     & n/a   & 76 \\
\bottomrule
\end{tabular}
\caption{\textbf{Demonstration parameters and results.}  
Parameters varied across the six demonstration cases in Fig.~\ref{demo}(a)--(f), showing effects of chamber pressure, actuator pressure, tip load, and fabrication style on achieved height and curve angle.}
\label{table demo}
\vspace{-2.0em}
\end{table*}

Here, we present a demonstration (Fig.~\ref{demo}) of how these recommendations make it possible to design and control two new robots to succeed in tasks where a robot with uninformed design and control parameters failed. Each robot performed two tasks that are critical for exploration of field environments: lifting the robot tip over an elevation change, and performing a high curvature turn. For the first task, each robot raised its tip as high as possible under its given load and chosen parameters. This maximum reach is represented by an elevated green archway that the robot could grow through. For the second task, the robots grew a certain distance and then curved to their maximum planar curvature, again represented by a green archway that they could then access. The results of these trials are shown in Fig.~\ref{demo}, with (a), (c), and (e) showing the elevation change results and (b), (d), and (f) showing the planar curve results. The parameters chosen for each robot, along with each robot's performance, are shown in Table~\ref{table demo}. All pressures were adjusted in increments of 3.45~kPa for simplicity.

The initial robot design representing the case where the design and control parameters were uninformed was inspired by the design presented in \cite{mcfarlandSPROUT}, which struggled to steer in the field while carrying its tip-mounted camera. That robot had a tip load of 145~g, maximum chamber and actuator pressures of 10.34~kPa, a body diameter of 8~cm, and exterior actuators. That tip load was the lowest amount that supported the camera design, whereas the pressure maximums were based on a similar design previously deployed in the field \cite{CoadRAM2020}. To mimic this robot, we used a design with an 8~cm body diameter and exterior actuators, with both the chamber and actuators having pressure maximums of 10.34~kPa. The tip mount was a plastic sleeve with 150~g attached to it for simplicity. When operating at its maximum chamber and actuator pressure, as was done in \cite{mcfarlandSPROUT}, it had a pressure ratio of 1. Under these conditions, it could only lift its tip over a 2~cm obstacle (Fig.~\ref{demo}(a)). For the growing task, the robot grew 21.3~cm, denoted by the yellow tape marks shown in Fig.~\ref{demo}, before curving in the plane. It required 10.34~kPa to grow. It was able to steer approximately 22° as estimated from the video (Fig.~\ref{demo}(b)).     

For the first informed design, we chose to keep the same body diameter and exterior pouches, but change the tip load and pressure ratio. This robot's chamber and actuators had a maximum pressure of 13.79~kPa. The tip load was reduced to 100~g to realistically represent the weight of sensors that might be needed for search and rescue tasks, while also being mindful of the impact of weight shown in Fig.~\ref{key experiments}(a). When operating with a chamber pressure of 6.89~kPa and a pressure in the top two actuators of 13.79~kPa, this pressure ratio of 2 allowed the robot to lift its tip over a 12~cm obstacle (Fig.~\ref{demo}(c)). It was necessary to have 6.89~kPa in the chamber to make the robot stiff enough to support its weight. For the growing task, the robot required 10.34~kPa to grow. However, the chamber pressure was dropped to 3.45~kPa for the planar curve, and the actuator pressure was set to 13.79~kPa. In this case, the robot did not need to support its own weight, so we could use a higher pressure ratio to perform the curve. This pressure ratio of 4 enabled the robot to steer 68° as estimated from the video (Fig.~\ref{demo}(d)).  

For our second informed design, we chose to keep the same body diameter and tip load from our first informed design, but we switched to a robot with integrated pouches. This robot's chamber and actuators also had a maximum pressure of 13.79~kPa. When operating the chamber at 6.89~kPa and the top two actuators at 13.79~kPa, a pressure ratio of 2, the robot was only able to lift its tip over a 2~cm obstacle (Fig.~\ref{demo}(e)). 
% However, it also steered to the side slightly instead of straight up, which may have prevented the force from pulling it to its highest possible curvature. 
It was necessary to keep the chamber at 6.89~kPa to improve the robot's ability to hold up its weight. For the growing task, this robot was able to grow at just 6.89~kPa. For the planar turn, the chamber pressure was reduced to 3.45~kPa, as the robot was fully supported by the table. The actuator was set to 13.79~kPa, creating a pressure ratio of 4. It was able to steer 76° as estimated from the video, the tightest planar curve achieved by any robot in this trial (Fig.~\ref{demo}(f)). 

The robot with uninformed parameters was not able to lift its tip well or perform a sharp planar curve. One alternative design that lightened the payload and used a larger pressure ratio made it possible to lift the tip higher and steer better in the plane. Another alternative design with a lightened payload, larger pressure ratio, and different fabrication style did not improve tip lift, but produced significantly sharper planar steering than either other design. 
% This makes it clear that there is no one right design for a vine robot. Certain changes lead to improvements, but there are also behavioral trade-offs. 
This highlights the idea that design and control parameters should be chosen with the task in mind. Given the current restrictions on how much pressure these robots hold, exterior actuators are better for tasks where lifting is critical, while integrated pouches are better for tasks where the robot is fully supported by the ground and simply needs to steer.

\section{Conclusions}

In this paper, we experimentally studied how design and control parameters affect vine robot steerability in both free space and ground-supported scenarios. We tested variations in tip load, chamber pressure, length, diameter, body fabrication, and actuator pressure, and demonstrated their impact through comparative tasks with different robot designs. Our results show that steerability decreases with increasing tip load and increases with moderate chamber pressure. Longer robots generally steer better, though excessive length can lead to collapse in free space, while diameter has minimal effect once sufficient to resist collapse. Regarding fabrication methods, exterior pouch actuators begin curving the robot at lower ratios of actuator to chamber pressure but offer limited maximum curvature, whereas integrated pouch actuators demand higher ratios yet achieve greater maximum curvature. Together, these findings provide experimentally grounded guidelines for tailoring vine robot design and control to field tasks, highlighting trade-offs between payload, pressure management, and actuator design.

In future work, we will explore robot body designs that exhibit less hysteresis in steering, as well as those that use proprioceptive sensing to detect their own shape and relay it to the human operator. We will also explore the development of improved actuator designs that sustain higher pressures while remaining compatible with vine robot growth. 
% This could involve alternative actuator geometries, reinforcement strategies, or new fabrication materials. 
Finally, we will explore vine robot control with time-varying actuator to chamber pressure ratios to enable selectively faster growth or greater steering depending on current operator needs.

\bibliographystyle{IEEEtran}
\bibliography{library}
\begin{biography}[{\includegraphics[width=1in,height=1.25in,clip,keepaspectratio]{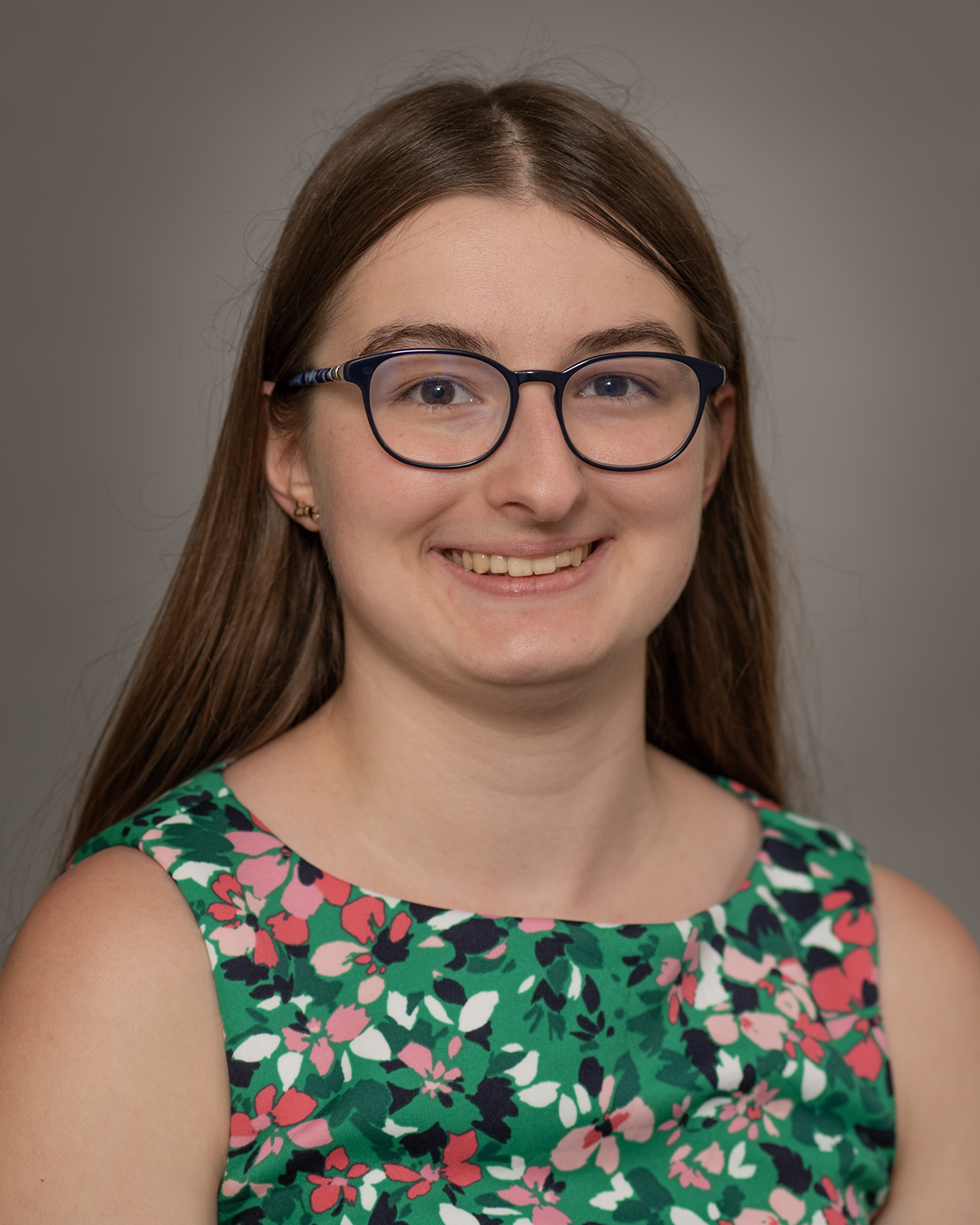}}]{Ciera McFarland} (Student Member, IEEE) received her B.S. degree in Aerospace Engineering from Pennsylvania State University and her M.S. degree in Mechanical Engineering from the University of Notre Dame. She is currently a Ph.D. candidate in Aerospace and Mechanical Engineering at the University of Notre Dame. Her research interests include soft robotics, disaster robotics, and human-robot interaction.
\end{biography}
\begin{biography}[{\includegraphics[width=1in,height=1.25in,clip,keepaspectratio]{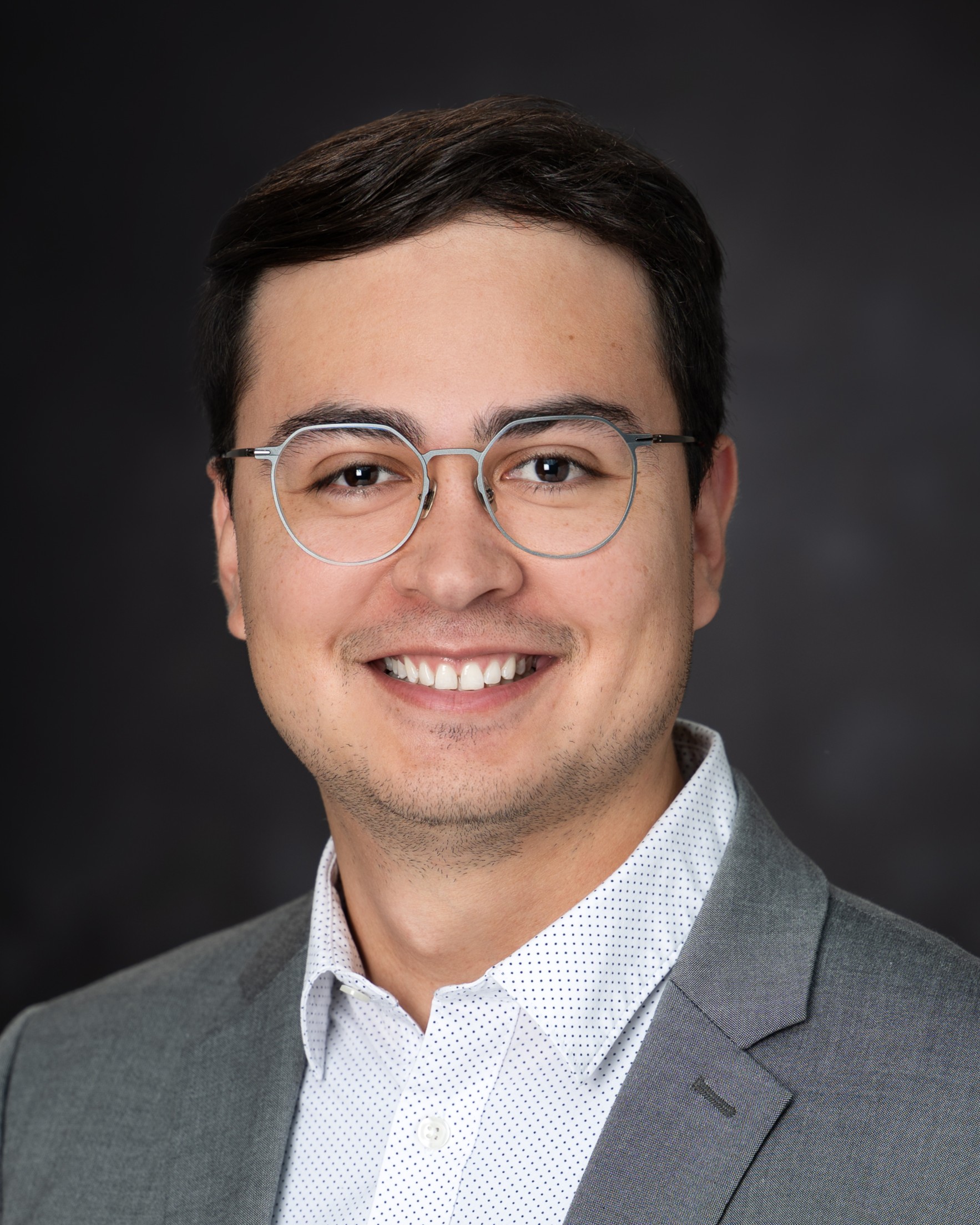}}]{Antonio Alvarez Valdivia}(Student Member, IEEE) received the B.S. degree in Mechanical Engineering from Iowa State University, and the Ph.D. degree in Mechanical Engineering from Purdue University. His interests include soft haptics, soft robotics, human-machine interaction, and mechatronics.
\end{biography}
\begin{biography}[{\includegraphics[width=1in,height=1.25in,clip, keepaspectratio]{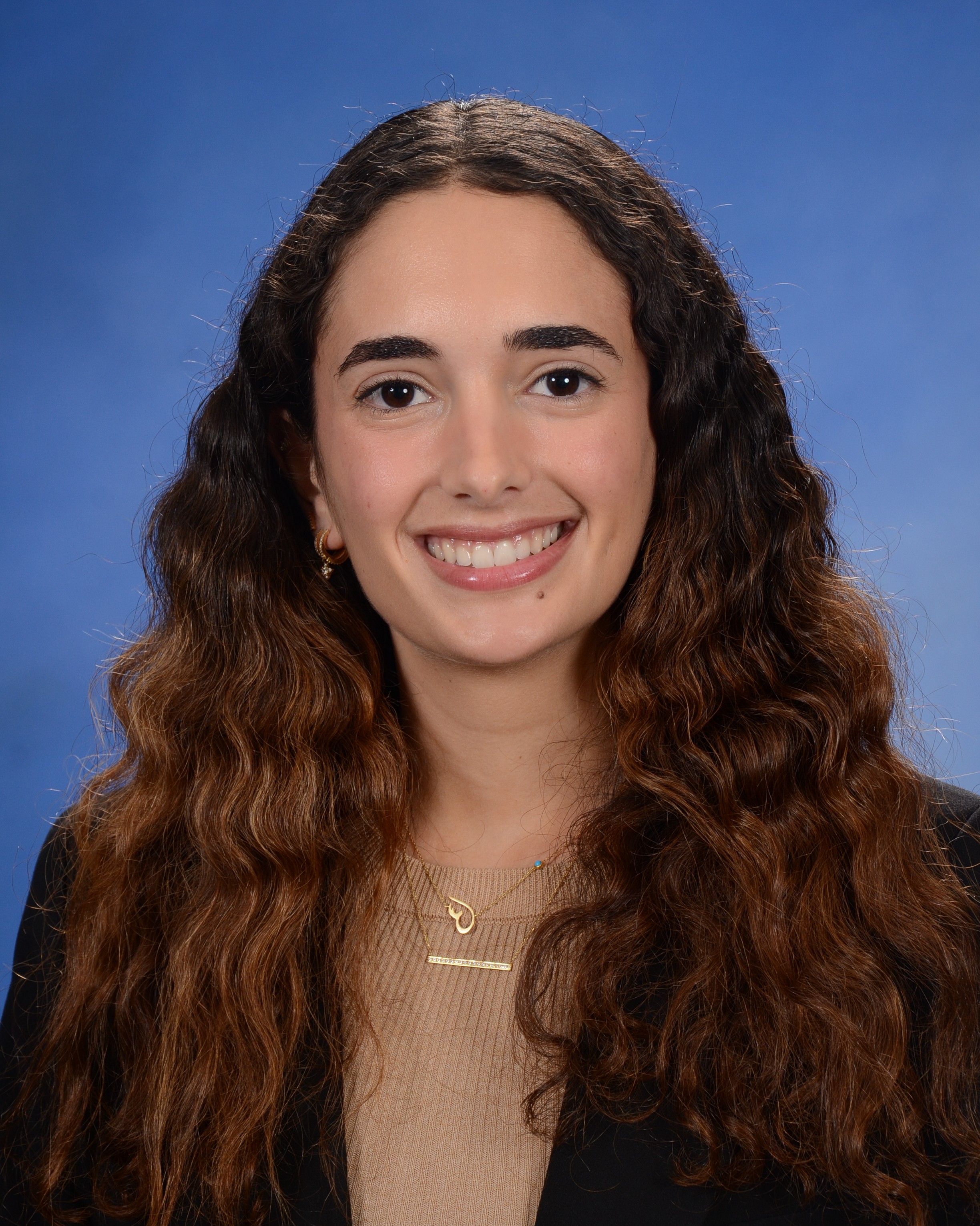}}]{Sarah Taher}(Student Member, IEEE) received her B.S. degree in Mechanical Engineering from the University of Notre Dame. During her undergraduate studies, she worked in the Innovative Robotics and Interactive Systems (IRIS) Lab, where she conducted research on vine robots for urban search and rescue. She is currently a Junior Sustainability Designer at Skidmore, Owings and Merrill (SOM).
\end{biography}
\begin{biography}[{\includegraphics[width=1in,height=1.25in,clip]{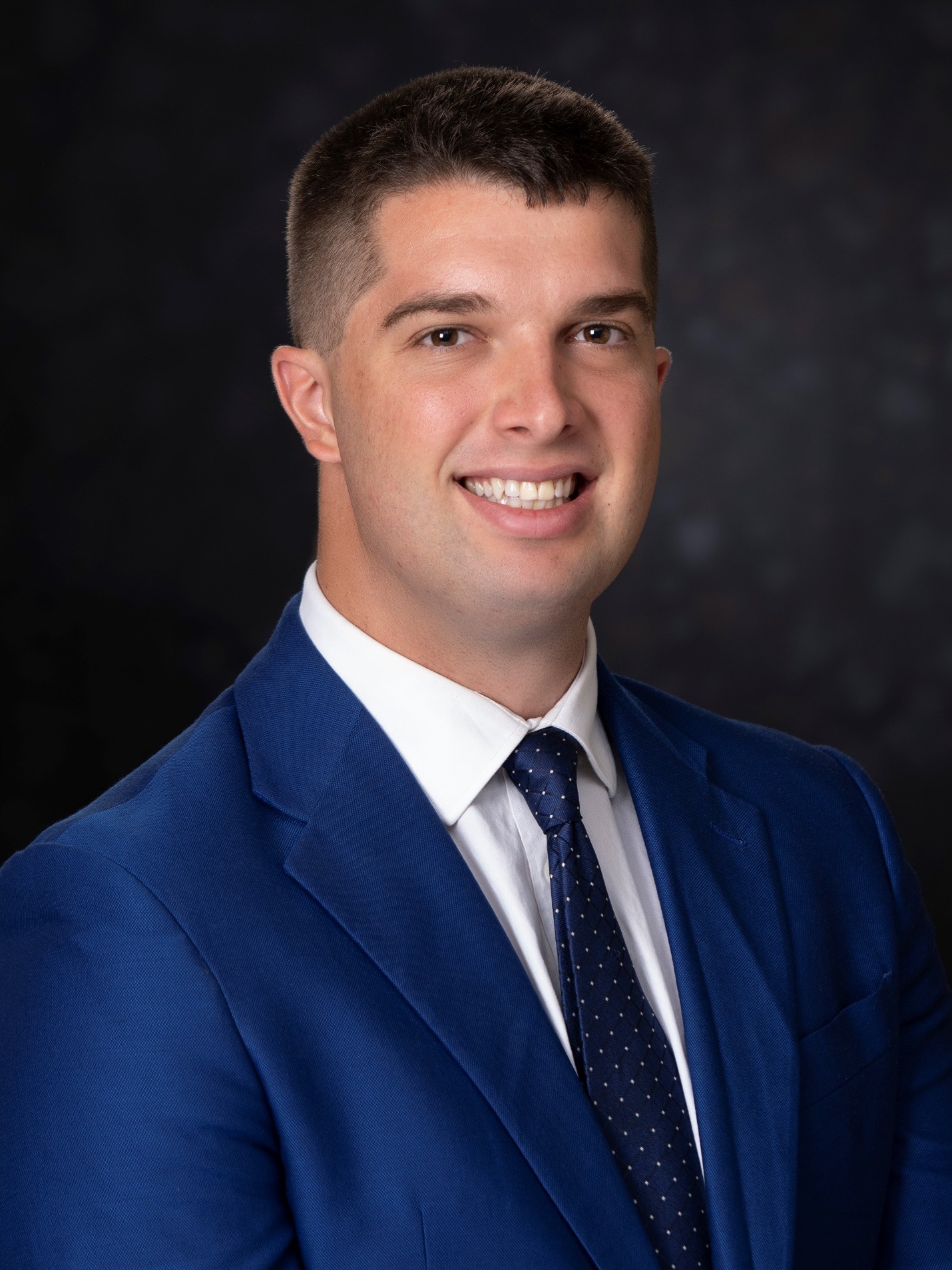}}]{Nathaniel Hanson} (Member, IEEE)
received his B.S. degree in Computer Engineering from the University of Notre Dame, M.S. degree in Computer Science from Boston University, and Ph.D. degree in Computer Engineering from Northeastern University. He is a member of the technical staff at the Massachusetts Institute of Technology Lincoln Laboratory in the Humanitarian Resilience Technology group. His research interests include disaster robotics, soft robotics, and sensor fusion.
\end{biography}
\begin{biography}[{\includegraphics[width=1in,height=1.25in,clip,keepaspectratio]{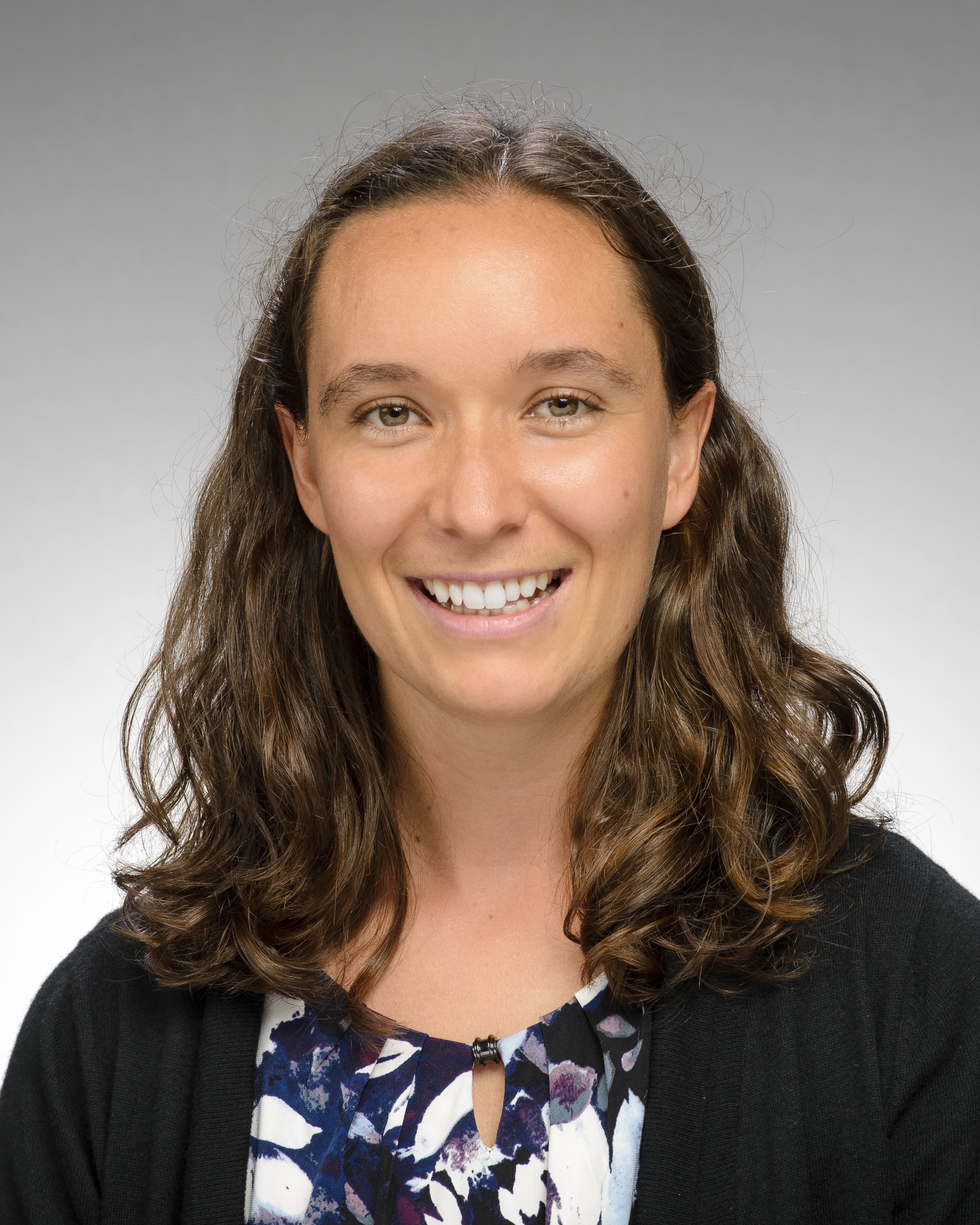}}]{Margaret McGuinness} (Member, IEEE) received her B.S. degree from Massachusetts Institute of Technology, and her M.S. and Ph.D. degrees from Stanford University, all in Mechanical Engineering. She is an Assistant Professor of Aerospace and Mechanical Engineering at the University of Notre Dame, where she leads the Innovative Robotics and Interactive Systems Lab. Her research interests include soft robotics, robot design and modeling, human-in-the-loop control, and field robotics.
\end{biography}
\end{document}